\documentclass[conference,twoside]{IEEEtran}
\IEEEoverridecommandlockouts
\usepackage[a4paper,top=0.825in,bottom=1.25in,right=0.68in,left=0.68in]{geometry}
\usepackage{fancyhdr}
\ifCLASSINFOpdf
\else
\fi
\usepackage{graphics}
\usepackage{subfigure}
\usepackage{epsfig}
\usepackage{multirow,rotating}
\usepackage{amsmath}
\usepackage{array}
\usepackage{multirow}
\usepackage{colortbl}
\usepackage{cite}
\usepackage{url}
\usepackage{times}
\usepackage{amssymb}
\usepackage{subfigure}
\usepackage{subfig}
\usepackage{tabularx}
\usepackage{algorithm,algpseudocode}
\usepackage{mathtools}

\usepackage{booktabs}

\usepackage{algpseudocode}
\usepackage{amsmath}
\usepackage{amsfonts}
\renewcommand{\exp}[1]{{\mathbb E}\left[ #1 \right]}
\begin{document}
\title{Video Synopsis Generation Using Spatio-Temporal Groups}


\author{
\IEEEauthorblockN{A. Ahmed and S. Kar}
\IEEEauthorblockA{NIT Durgapur, India\\
 Email: arif.1984.in@ieee.org,\\
 samarjit.kar@maths.nitdgp.ac.in}

\and \IEEEauthorblockN{D. P. Dogra}
\IEEEauthorblockA{IIT Bhubaneswar, India\\
 Email: dpdogra@iitbbs.ac.in}\\

\IEEEauthorblockN{H. Choi and I. Kim}
\IEEEauthorblockA{KIST, Seoul, Republic of Korea\\
 Email:\{hschoi, drjay\}@kist.re.kr}

\and \IEEEauthorblockN{R. Patnaik and S. Lee}
\IEEEauthorblockA{IKST Bangalore, India\\
 Email: \{renuka.patnaik,seungcheol.lee\}@ikst.res.in}

 }


%Telephone: (800) 555--1212, Fax: (888) 555--1212}


\maketitle
%\renewcommand{\headrulewidth}{0pt}
%\renewcommand{\footrulewidth}{0pt}
%\thispagestyle{fancy} \lfoot{xxx \copyright 2017 IEEE} \cfoot{}
% \pagestyle{fancy}
%\fancyhead[LO,RE]{{\small 2017 IEEE International Conference xxxx}}

\begin{abstract}
Millions of surveillance cameras operate at 24x7 generating huge
amount of visual data for processing. However, retrieval of
important activities from  such a large data can be time
consuming. Thus, researchers are working on finding solutions to
present hours of visual data in a compressed,  but meaningful way.
Video synopsis is one of the ways to represent activities using
relatively shorter duration clips. So far, two main approaches
have been used by researchers to address this problem, namely
synopsis by tracking moving objects and synopsis by clustering
moving objects. Synopses outputs, mainly depend on tracking,
segmenting, and shifting of moving objects temporally as well as
spatially. In many situations, tracking fails, thus  produces
multiple trajectories of the same object. Due to this, the object
may appear and disappear multiple times within the same synopsis
output, which is misleading.  This also leads to discontinuity and
often can be confusing to the viewer of the synopsis. In this
paper, we present a new approach for generating compressed video
synopsis by grouping tracklets of moving objects. Grouping helps
to generate a synopsis where chronologically related objects
appear together with meaningful spatio-temporal relation. Our
proposed method produces continuous, but a less confusing synopses
when tested on publicly available dataset videos as well as
in-house dataset videos.
\end{abstract}


% \fi
\IEEEpeerreviewmaketitle

\section{Introduction}
Surveillance videos recorded at 24x7 are normally not of much use
unless summarized meaningfully. There exist a few approaches to
produce video synopses. For example, the fast-forward approach
proposed in~\cite{wildemuth2003fast} is well-known to generate
video synopsis. Unfortunately, the method misses events such as
fast moving objects while skipping video frames. To mitigate this
problem, a few alternatives have been proposed
in~\cite{petrovic2005adaptive,ma2002model}, where key frames are
chosen conditionally to generate synopsis. However, frame-based
approaches tend to produce longer duration synopsis by combining
activity video clips
sequentially~\cite{pritch2008nonchronological,nie2013compact}. To
reduce the length of the synopses, a number of approaches have
been proposed that extract activity area from the raw video and
montage these activities
together~\cite{pal2005interactive,manickam2013automontage}. In the
synopsis video, several activities coming from different times are
stitched in continuous frames. In this case, it often produces
unpleasant synopsis due to blending seams coming from different
image patches.

To address such problems, researchers have proposed object-based
non-chronological video synopsis
approach~\cite{pritch2008nonchronological,rav2006making,pritch2009clustered}.
In object-based approaches, moving objects are tracked using
multi-object tracker, segmented at real-time, and then shifted to
different time in the synopsis according to user's requirements.
Although object-based methods can reduce duration of synopsis, it
may cause collisions between objects that appear within same
spatial domain and sometimes produces confusing synopsis by
showing several activities simultaneously. To reduce such
collisions, a few improvements  have already been proposed. For
example, Nie et al.~\cite{nie2013compact} and Kang et
al.~\cite{kang2006space} have proposed methods by shifting the
moving objects in spatial and temporal domain. Although these
methods reduce collisions, both temporal and location information
of objects are violated. Pritch et
al.~\cite{pritch2008nonchronological} have proposed a method by
energy minimization with only temporal shifting of objects.
Results produced by their method produces confusing synopsis as
several activities are shown simultaneously. Xuelong et
al.~\cite{li2016surveillance} have proposed methodology to
minimize object collision with minimum time shifting by scaling
down the object size. Pritch et al.~\cite{pritch2009clustered}
have proposed method by clustering moving objects and showing
similar activity together to deal with the confusing synopsis.
Their method deals with trajectory clustering and displays similar
activities together.

Video synopsis methods enable fast and efficient browsing of
surveillance videos, however create a summary that are often
confusing to humans for visual inspection. In this paper, we have
proposed a new method that is primarily built upon the concept
proposed
in~\cite{pritch2008nonchronological,pritch2009clustered,rav2006making,yao2014object}
and~\cite{li2016surveillance}. However, our proposed method
enables displaying a group of activities simultaneously which
originate from different time periods. Also, it creates meaningful
summaries by minimizing sudden appearance and disappearance of
objects, and also reduces confusions by grouping different
activities.

\subsection{Motivation and Contributions}
\begin{figure}[!ht]
\centering
\includegraphics[width=3.3in]{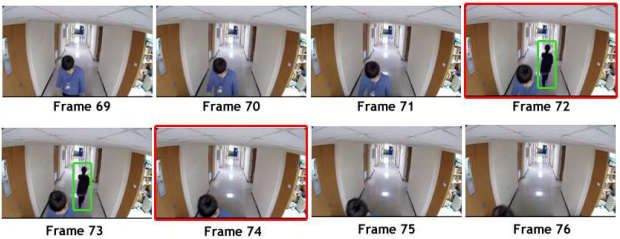}
\caption{Examples of two moving objects appear at different times
in the in-house KIST dataset video and how they appear in the
synopsis video. It has been observed that a small tube (activity
segment) appears in a very short period, For example, a sample
small tube appeared and disappeared in between Frame 72 to Frame
73 (red border). It creates sudden appearance and disappearance of
moving objects in the synopsis video. } \label{fig:disapp}
\end{figure}

\begin{figure}[!ht]
\centering
\includegraphics[width=3.3in]{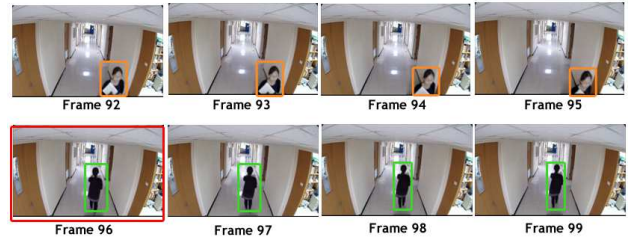}
\caption{Example of a synopsis segment (Frame 92 to Frame 99) of a
moving object that appears at different times in the KIST video.
It has been observed that, two small tubes or activities (green
and orange) appear non-chronologically. Tube represented by the
top row actually appears after the tube represented by the bottom
row in the original video. It also creates sudden appearance and
disappearance of objects, therefore does not reflect the ground
truth.} \label{fig:contt}
\end{figure}

\begin{figure}[!ht]
\centering
\includegraphics[width=3.3in]{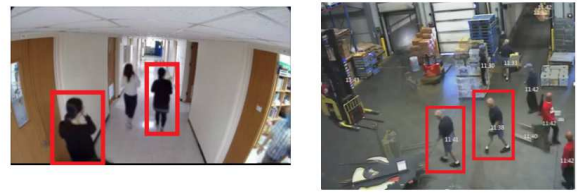}
\caption{Example of a set of synopsis frames taken from the KIST
video and the results synopsis reported
in-\cite{pritch2008nonchronological}. It has been observed that,
same object may appear multiple times at the same time frame.
Therefore, it creates confusing synopsis.} \label{fig:contt2}
\end{figure}

Short duration tube or activity based synopsis generation
described in
~\cite{pritch2008nonchronological,pritch2009clustered,rav2006making,yao2014object}
and~\cite{li2016surveillance} often produces poor synopsis. It can
produce synopsis with shorter lengths by showing simultaneously
multiple activity segment of same object, which is confusing.
Therefore, the outputs are not always meaningful to correlate with
ground truths by human video analysts. The algorithm suffers from
the below drawbacks:
\begin{itemize}
\item Sometimes objects may appear and disappear in the synopsis
video as depicted in Fig.~\ref{fig:disapp} and
Fig.~\ref{fig:contt}. \item Activity segments or tubes of the same
object may appear non-chronologically resulting discontinuous
synopsis as depicted in Fig.~\ref{fig:contt}. \item Activity
segments or tubes of the same object may appear at the same time
resulting confusing synopsis as depicted in Fig.~\ref{fig:contt2}.
\end{itemize}

This paper proposes a method to generate synopsis of a time
bounded video by overcoming aforementioned drawbacks. In contrast
to these methods proposed
in~\cite{pritch2008nonchronological,pritch2009clustered,rav2006making,yao2014object,li2016surveillance},
our proposed method groups the moving objects to generate the
synopsis. Grouping is done with respect to relative
spatio-temporal distance and chronological appearance of the
objects. Results reveal the superiority of our method over
existing approaches. Also, our method produces synopsis that
originally reflect continuous activities of moving objects. For
example, An object appeared in a scene at 10:00:00 AM and exit the
scene at 10:01:15 AM. Proposed method preserve the chronological
appearance of the same object in different location, and produce
continuous synopsis of the same object.

Rest of the paper is organized as follows. In the next section, we
present the proposed methodology in detail.
Section~\ref{sec:results}, we present the results obtained using
publicly available datasets as well as using the videos of
in-house KIST dataset. Conclusion and future directions of the
present work are discussed in Section~\ref{sec:conclusion}.

\section{Proposed Synopsis Generation Method}
\label{sec:proposed} 
\begin{figure}[!ht]
\centering
\includegraphics[width=3.3in]{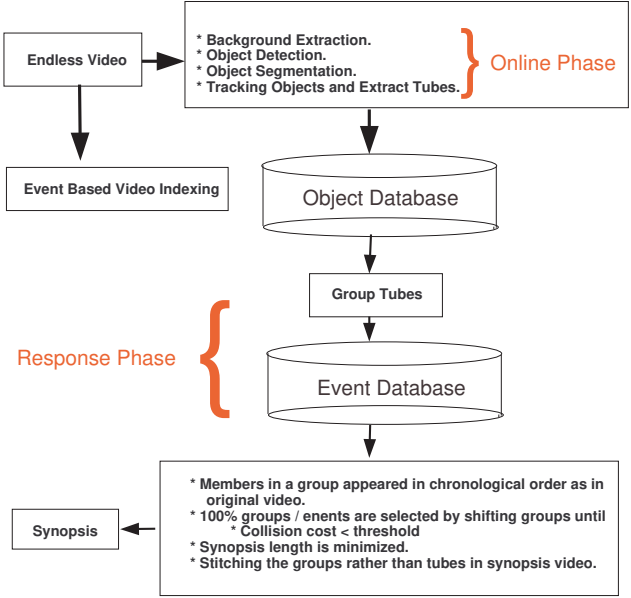}
\caption{Our proposed method of generating synopsis of a
time-bound video. In real-time, moving objects are tracked and
segmented, known as tubes. Tube database is actually populated
during video recording. Synopsis is generated in response phase by
grouping and optimizing the length.} \label{fig:prop}
\end{figure}

In this section, we will present the proposed
method in detail. The main approach is presented in
Fig.~\ref{fig:prop}. The method can be divided into two phases
similar with the method proposed
in~\cite{pritch2008nonchronological}. First, moving objects are
segmented using improved Mixture of
Gaussian~\cite{zivkovic2004improved} and then tracked using Kalman
filter based multi-object tracker~\cite{welch2001introduction}.

A tube/activity is defined by the continuous position of the
object, and activity segment is the part of the scene between
entry and exit in a scene of an object. Tracking and extracting
activity is done in real-time which is denoted by the
online-phase. Next, moving objects are grouped in unsupervised
manner and then each group is stitched into the background using
the Poisson image blending~\cite{perez2003poisson}. Stitching is
done by blending multiple activity in a known background. Synopsis
length $(L)$ is also minimized before generating the synopsis
during this response phase. A response phase is responsible to
generate the synopsis of a time bounded video according to user
requirements.

Video synopsis can be considered as an index of the original
video. In addition, video synopsis can save storage for storing
surveillance videos by discarding frames when there is no
activity. The quality of the synopsis highly depends on the
content of the original video and the length of the synopsis.
Also, there are lagging of standard quantitative measurement that
method can be used to compare the quality of synopsis. However,
there are some baseline standards in video
synopsis~\cite{pritch2008nonchronological,pritch2009clustered}.
They are as follows:

\begin{itemize}
 \item  The video synopsis should be substantially shorter than the original video and should preserve the maximum activities
presented in the original video.
 \item Collision among objects, i.e. overlapping should be minimized and if possible, be avoided to produce smooth synopsis.
 \item Temporal relation among objects, i.e. interaction among objects must be preserved in the final synopsis.
\end{itemize}

To maintain quality of the synopsis as mentioned earlier, we
present a method for grouping object activities or tubes before
generating the final synopsis. We calculate the energy difference
between original and synopsis videos. Energy is defined by
calculating pairwise spatio-temporal relation and chronological
order of appearance among the objects.

\subsection{Energy Differences}
We first define the energy difference between original and
synopses video. If the difference is lower, the synopsis quality
will be higher. The energy summarizes interaction and
chronological appearance among segment of objects. Let $O$ be the
original video and $S$ be its synopsis. Each tube  $t$ is defined
over a time-bound segment in the original video stream
$t_b=[{t_b}^s,{t_b}^e]$, where ${t_b}^s$ and ${t_b}^e$ are the
start and end frames.

The synopsis video ($S$) is generated based on a temporal mapping
$M$ over $O$. The mapping defines the shifting of objects $b$ into
the time segment $t_{\hat{b}}=[{t_{\hat{b}}}^s,{t_{\hat{b}}}^e]$
in the synopsis video. $M(b)=\hat{b}$ indicates the time shift of
tubes $b$ into the synopsis. Optimized synopsis of the video is
generated by minimizing the following energy function defined by
modifying the energy function originally presented
in~~\cite{pritch2008nonchronological}. The energy is
defined~(\ref{eq:energy1}), where $b$ and $b'$ represent the tubes
present in the video
\begin{small}
\begin{equation}
\label{eq:energy1} E(M)= \sum_{b, b'\in{S}}{ \big (
E_a(\bar{b}\cup \bar{b'})+ E_t(b,b') + E_o(b,b')+ E_c(b,b' )} \big
)
\end{equation}
\end{small}

Activity cost of objects is defined by the continuous position of
the object during a bounded time. $E_a$ is the activity cost of
the tube segment $(\bar{b}\cup \bar{b'})$ that is not included in
the synopsis, $E_t$ is the spatio-temporal consistency cost, $E_o$
is the chronological appearance cost and $E_c$ is the collision
cost. Higher collision cost $(E_c)$, for example, will result in a
denser video, where objects may overlap.

When collision cost and activity cost in the final synopsis are
zero, i.e. $\sum{E_c}=0$ and $\sum{E_a(\bar{b}\cup \bar{b'})}=0$,
all activity tubes are mapped in the synopsis video. Therefore,
the energy of the synopsis can be defined
using~(\ref{eq:energy2}),  where the spatio-temporal consistency
cost $(E_t(b,b'))$ is the spatio-temporal distance between $b$ and
$b'$ in original and synopsis videos.

\begin{equation}
\label{eq:energy2} E(M)= \sum_{b,b'\in{S}}{\big ( E_t(b,b') +
E_o(b,b')\big )}
\end{equation}

$E_t(a,b)$ is defined in~(\ref{eq:energy3}), where ${b,b'}\in{O}$
and $\hat{b},\hat{b'}\in{S}$.
\begin{eqnarray}
\label{eq:energy3} E_t(b,b')= d(b,b')-d(\hat{b},\hat{b'}),
\end{eqnarray}

When an object is shifted from a group of interacting objects,
cost of the synopsis increases. The distance $(d(a,b))$ can be
defined as given in~(\ref{eq:energy4}), where $t_a$, $t_b$
represent the time of appearance of $a$ and $b$.
\begin{equation}
\label{eq:energy4}  d(a,b)=\begin{cases}
    0, & \text{if $t_{a} \bigcap t_{b}= \phi$}.\\
    d_s(a,b), & \text{otherwise}.
  \end{cases}
\end{equation}

 $t_a \bigcap t_{b}$ is the time intersection of $a$ and $b$ in the video. The amount of interaction $d(a,b)$ between each pair of tubes is calculated from their relative spatio-temporal distance, as defined in~(\ref{eq:energy5}),  where $d(a,b,t)$ is the Euclidean distance between the tubes at time $t$, and $\sigma_{area}$ is the area of the object.
\begin{equation}
\label{eq:energy5}  d_s(a,b)= \exp{ \min_{t\in{t_{a}\bigcap
t_{b}}} \big \{ d(a,b,t) / \sigma_{area} \big \} }
\end{equation}

Chronological appearance cost ($E_o$) is the cost of violating
chronological order in the synopsis. $E_o$ is defined in
~(\ref{eq:energy6}). When objects appear in the synopsis by
violating their relative chronological order, the cost becomes
higher.
\begin{equation}
\label{eq:energy6}
 E_o(b,b')= d(b,b')-d(\hat{b},\hat{b'})
\end{equation}

The cost is calculated using~(\ref{eq:energy7}), where $d_{ch}$ is
the constant cost for violating order of activity appearing in the
synopsis, ${b,b'}\in{O}$, and ${\hat{b}},\hat{b'}\in{S}$.
\begin{equation}
\label{eq:energy7} d(a,b)= \big( t_{a}^s - t_{b}^s \big).
\begin{cases}
    0, & \text{if $t_{a} \bigcap t_{b}\neq\phi$}.\\
    d_{ch}(a,b), & \text{otherwise}.
  \end{cases}
\end{equation}
The cost is further refined using~(\ref{eq:energy8}), where the
constant cost $C$ is chosen as one.
\begin{equation}
\label{eq:energy8} d_{ch}(a,b)=\begin{cases}
    0, & \text{if $t_{b}^s - t_{b'}^s=t_{\hat{b}}^s - t_{\hat{b}'}^s$}.\\
    C, & \text{otherwise}.
  \end{cases}
\end{equation}

\subsection{Effect of Energy Difference $E(M)$}
The length of the synopsis $(L)$ depends on the energy difference
$E(M)$. If $E(M)=0$, there is no difference between original video
and synopsis. Hence $L$ is equal to the length of the original
video. If $E(M)$ is maximum, we allow a maximum shifting of tubes.
It may produce a synopsis of different length based on the
original content of the video. However, sometimes tracking of
objects fails loosing identity~\cite{yang2011recent} or force
fragmenting tubes as described
in~\cite{pritch2008nonchronological}. It creates multiple tubes
for the same moving object or group of objects. During
optimization, these tubes are considered as different objects.
Generated synopsis using these short-duration tubes often suffers
from the below drawbacks:
\begin{itemize}
\item Shifting tubes randomly may produce confusing synopsis by
showing multiple appearances of an object in the same frame. \item
Sometimes, tubes belonging to the similar activity group violate
the order of chronological appearance in synopsis. It may cause
sudden appearance and disappearance of the objects and loss in
chronological activity may be observed in the synopsis. This is
explained as follows. Let, $t_a$ and $t_b$ be two tubes such that
$t_a^s<t_b^s$. In the synopsis, if $t_a$ and $t_b$ are part of the
same moving object, i.e. $t_a,t_b \in{T}$,  when $t_b$ appears
before $t_a$, it loses the chronological ordering. \item
Sometimes, objects are shifted to different temporal segment to
minimize the synopsis length despite having strong interaction.
For example, they may share similar temporal segment in the
original video. It loses the interaction information.
\end{itemize}

\subsection{Grouping of Tubes and Synopsis Length Minimization}
To overcome the aforementioned problems, we have proposed a method
of grouping object trajectories or tubes. Shifting of objects in
the synopsis is then restricted by grouping them together.
Relative spatio-temporal distance and chronological appearance of
objects in the same group is unchanged in the final synopsis.
Hence $E_t(a,b)=0$ and $E_o(a,b)=0$ such that $a,b\in{G}$.
Grouping helps to bind related tubes together. Groups are
generated based on the spatio-temporal distance $(d_s(a,b))$ and
chronological distance. Let the original video $(O)$ be
represented using~(\ref{eq:scene}), where $t_i$ represents the
tubes present in the scene. Tubes are grouped together, and the
scene is represented using~(\ref{eq:scene2}), where $G$ is a set
of tubes grouped together.
\begin{equation}
\label{eq:scene} O=\{t_1,t_2,...,t_n\}
\end{equation}
\begin{equation}
\label{eq:scene2} S=\{G_1,G_2,...,G_m\}
\end{equation}

$t_a$ and $t_b$ are assumed to be in the same group, if they
chronologically appear or interact within a fixed threshold. It
can be expressed as $t_a,t_b \in G$, when $d_s(a,b)<\alpha$ or
$t_{a}^s - t_{b}^s<\beta$, where $\alpha$ is the maximum distance
to measure the interaction and $\beta$ represents the maximum
chronological distance for grouping. The algorithm for grouping
the tubes are presented in Algorithm~\ref{alg:group}.

\begin{algorithm}
\caption{Grouping of Tubes} \label{alg:group}
\begin{algorithmic}[1]
\Procedure{GROUP TUBES}{$O$}\Comment{Group tubes}
 \State $R=\{\phi\}$
 \For{\texttt{(a=$t_1...t_n$)}}
    \If{$a\notin{R}$}
      \State $G=\{a\}$
      \State $R=R\cup G$
   \EndIf
       \For{\texttt{(b=$t_1...t_n$)}}
        \If{$b\notin{R}$}
            \If{$d_s(a,b)<\alpha$ OR $(t_{a}^s - t_{b}^s)<\beta$}
                \State $G=G\cup b$, where $a\in G$
                \State \textbf{break}
            \EndIf

        \EndIf
        \EndFor
 \EndFor
\State \textbf{return R} \EndProcedure
\end{algorithmic}
\end{algorithm}

\begin{figure}[!ht]
\centering
\includegraphics[width=3.0in]{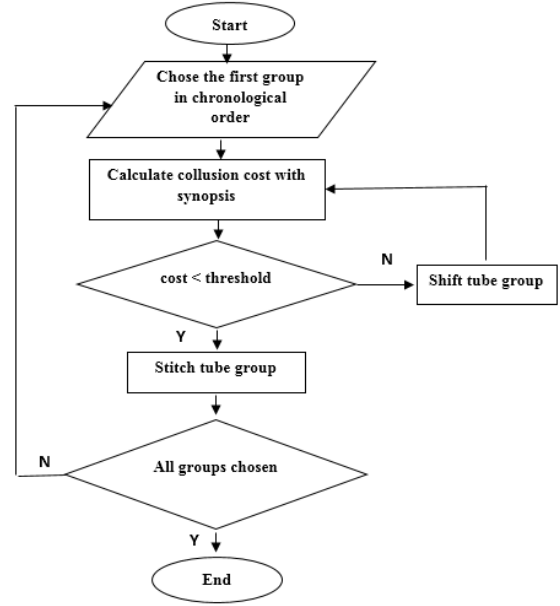}
\caption{Process of minimization of synopsis length.}
\label{fig:synop}
\end{figure}

The original video is represented by a set of groups. The synopsis
length is minimized by shifting and stitching the groups in the
synopsis. The groups are initially selected in chronological
order. Then, a group is fitted into a desired location. The
process is depicted in Fig.~\ref{fig:synop}.

\subsection{Effect of $\alpha$ and $\beta$}
The energy difference ($E(M)$) and length of synopsis ($L$) are
related to each other. Spatio-temporal threshold ($\alpha$) can be
defined in the boundary as
$\alpha=[\min(d_s(a,b)),\max(d_s(a,b))]$ and chronological
ordering threshold ($\beta$) defined in the boundary as
$\beta=[\min(t_{a}^s - t_{b}^s),max(t_{a}^s - t_{b}^s)]$, $\forall
a,b\in O$. If $\alpha=min(d_s(a,b))$ and $\beta=min(t_{a}^s -
t_{b}^s)$, then the number of groups is equal to the number of
tubes present in $O$. It may produce a synopsis with higher $E(M)$
when objects are shifted to reduce the synopsis length. Similarly,
when $\alpha=\max(d_s(a,b))$ and $\beta=\max(t_{a}^s - t_{b}^s)$,
all tubes belong to a single group, hence $E(M)=0$ resulting the
original video and the synopsis as of same length. It has also
been observed that, grouping spatio-temporal or chronological
tubes together, sometimes produces smaller synopsis.
\begin{figure}[!ht]
\centering
\includegraphics[width=3.0in]{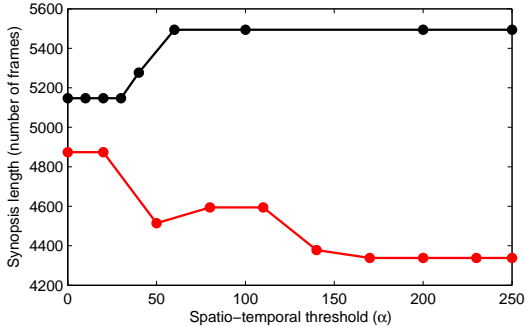}
\caption{Relation between spatio-temporal relation threshold
$(\alpha)$ and synopsis length. Red curve represents results using
KIST dataset and black curve represents results using VIRAT
dataset videos.} \label{fig:res1}
\end{figure}
\section{Experiments and Results}
\label{sec:results} In this section, we present qualitative and
quantitative analysis of our proposed method. We have experimented
with two datasets, namely VIRAT~\cite{oh2011large} and in-house
KIST. VIRAT dataset contains 16 minutes long video that are
publicly available and KIST is our in-house dataset of
approximately 30 minutes duration.

We first present the relation among spatio-temporal threshold
$(\alpha)$, chronological ordering threshold $(\beta)$, synopsis
length $(L)$, and energy difference $\big ( E(M) \big)$.
Fig.~\ref{fig:res1} shows the synopsis length by varying $\alpha$,
considering $\beta=0$. It has been found that the synopsis length
becomes almost constant beyond a certain value of $\alpha$. It
happens because all moving objects create a minimum number of
groups based on the $\alpha$.

\begin{figure}[!ht]
\centering
\includegraphics[width=3.0in]{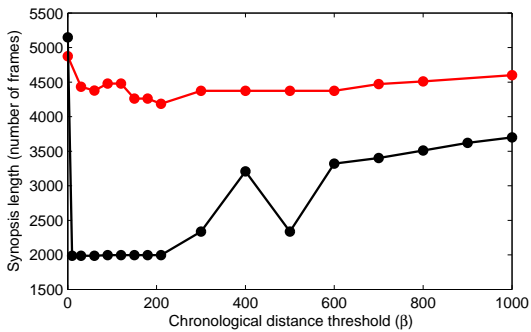}
\caption{Relation between chronological distance threshold
$(\beta)$ and synopsis length $(L)$. Red curve represents results
using KIST dataset and black curve represents results using VIRAT
dataset videos.} \label{fig:res2}
\end{figure}
Similarly, Fig.~\ref{fig:res2} shows the synopsis length by
varying $\beta$, considering $\alpha=0$. It has also been observed
that, synopsis length gradually increases after a certain value of
$\beta$. It happens because moving objects which appear
chronologically with large $\beta$ interval, are considered in the
same group. When $\beta$ reaches its  peak value, all moving
objects create a single group. The length of the synopsis then
becomes same as the original video.
\begin{figure}[!ht]
\centering
\includegraphics[width=3.0in]{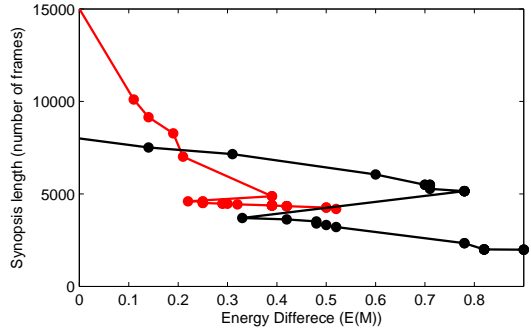}
\caption{Relation between synopsis length ($L$) and energy
difference $(E(M))$. Red curve represents results using KIST
dataset and black curve represents results using VIRAT dataset
videos.} \label{fig:res3}
\end{figure}

Fig.~\ref{fig:res3} shows how synopsis length vary when $\big
(E(M) \big)$ is varied. It has been observed that the energy
difference decreases when the length of the synopsis increases
after a threshold. The energy difference $(E(M))$ becomes zero
when the length of the synopsis is same as the length of the
original video.
\begin{figure*}[t]
\centering
\includegraphics[width=6.0in]{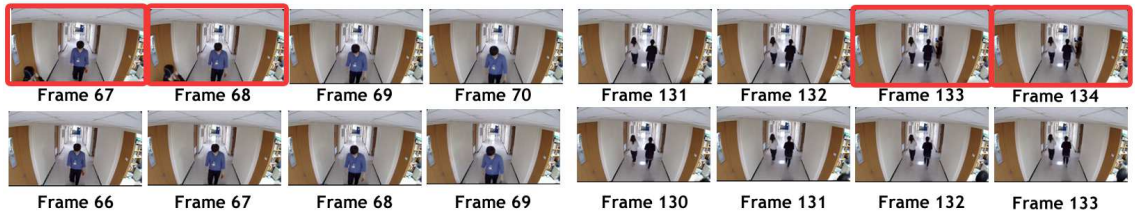}

\caption{Sample synopsis output frames taken from the KIST
dataset. First row represents synopsis when energy difference
$(E(M)$ is high, and the second row is the synopsis when energy
difference is low by grouping objects. It has been observed that,
sudden appearance and disappearance of small tubes (red bordered)
can be minimized when energy is low. It is visually less confusing
too.} \label{fig:snap_kist}
\end{figure*}

Fig.~\ref{fig:snap_kist} depicts a set of frames related to two
different synopsis outputs obtained by the algorithm proposed
in~\cite{pritch2008nonchronological,pritch2009clustered,rav2006making,yao2014object,li2016surveillance}
and using our proposed method. Frames that are marked red show
sudden appearance and disappearance of objects in the synopsis
generated using the method proposed
in~\cite{pritch2008nonchronological}. If the synopsis length is
minimized to show maximum activity, existing method produces
synopsis with high $E(M)$ as depicted in the first row. It has
been observed that, when $E(M)$ is low by grouping objects (second
row), it produces better synopsis outputs. It has been observed
that when the synopsis is generated
using~\cite{pritch2008nonchronological,pritch2009clustered,rav2006making,yao2014object,li2016surveillance},
due to high $E(M)$, the outputs are often discrete and confusing
as compared to our method.  We have restricted shifting of objects
by grouping, therefore,
 our method produces longer synopsis with a lower energy difference.

\section{Conclusion}
\label{sec:conclusion} We have proposed a method of video synopsis
generation. Our proposed method produces more meaningful synopsis
as compared to the existing methods. It has been observed that, by
grouping object trajectories, spatio-temporal relations among the
objects can be preserved with higher accuracy. We have tested our
methodology on publicly available video dataset and in-house
dataset. Initial results are encouraging and the method can be
applied on larger scale. There are many possible extensions of the
present work. We can group the trajectories based on various other
criteria such as interest area
based~\cite{ahmed2016localization,dogra2016smart,Dogra2015Interest},
movement graph based~\cite{Dogra2015Scene}, by supervised or
unsupervised machine
learning~\cite{saini2017surveillance,saini2017classification}, or
by using deep learning to understand the activities based on
region(s) of interest~\cite{ahmed2016localization}.

\section*{Acknowledgement}
The work has been funded under  Global Knowledge Platform (GKP)
institutional program scheme (Project No. 2E27190) of Indo-Korea
Science and Technology Center (IKST) and Korea Institute of
Science and Technology (KIST)  executed at IIT Bhubaneswar under
Project Code: CP106.

\bibliographystyle{IEEEtran}



%\vfill


\end{document}